\documentclass{llncs}
\usepackage{graphicx}
\usepackage{color}
\usepackage{amsmath}
\usepackage{amssymb}
\usepackage{xspace}
\usepackage{url}
\usepackage{hyperref}

\makeatletter
\DeclareRobustCommand\onedot{\futurelet\@let@token\@onedot}
\def\@onedot{\ifx\@let@token.\else.\null\fi\xspace}

\def\ie{\emph{i.e}\onedot}

\newcommand{\figcaption}[1]{\def\@captype{figure}\caption{#1}}
\newcommand{\tblcaption}[1]{\def\@captype{table}\caption{#1}}
\makeatother

\def\title#1{{\noindent\Large{\bf #1}\par}}
\def\author#1{\begin{center}{\sc #1\par}\end{center}}
\renewcommand{\section}[1]{\vspace{0.1in}\noindent{\large\bf{#1}}\par\vspace{.05in}\par\nopagebreak}

\makeatletter
\def\thebibliography#1{\section{References\@mkboth
{REFERENCES}{REFERENCES}}\list
{[\arabic{enumi}]}{\settowidth\labelwidth{[#1]}\leftmargin\labelwidth
\advance\leftmargin\labelsep
\usecounter{enumi}}
\def\newblock{\hskip .11em plus .33em minus .07em}
\sloppy\clubpenalty4000\widowpenalty4000
\sfcode`\.=1000\relax}
\makeatother

\begin{document}
\pagestyle{empty}

\title{Action recognition in real-world videos}
\author{Waqas Sultani, Information Technology University, Pakistan \\ Qazi Ammar Arshad, Information Technology University, Pakistan \\ Chen Chen, University of North Carolina at Charlotte, USA}

\section{Related Concepts}
\begin{itemize}
\item{Action recognition and localization}
\item{Action based video summarization}
\end{itemize}

\section{Definition}
The goal of human action recognition is to temporally or spatially localize the human action of interest in video sequences. Temporal localization (\ie indicating the start and end frames of the action in a video) is referred to as frame-level detection. Spatial localization, which is more challenging, means to identify the pixels within each action frame that correspond to the action. This setting is usually referred to as pixel-level detection.
In this chapter, we are using action, activity, event interchangeably.

\section{Background}

Three main ingredients of action research are visual features, machine learning methodology, and datasets. Recent years have witnessed a tremendous increase in research and development in all these areas of research. Several new visual features have been proposed which range from handcrafted local and global features and deeply learned visual features for action recognition and detection.  Almost all the machine learning techniques have been applied to achieve robust action classification. Most of the action classifications methods gear around supervised approach \cite{jain201515,Choutas_2018_CVPR,hou2019efficient}. Since obtaining labels of videos for the supervised approach is quite a time consuming and costly, several weakly supervised \cite{chang2019d,kumar2017hide,cvpr17wang,cvf18nguyen,zhong2019graph} and unsupervised approaches \cite{soomro2017unsupervised,yu2011unsupervised,jones2014multigraph} have been proposed.

The availability of diverse and real-world representative datasets plays a crucial role in research and development in any field.
Several large scales, diverse, real-world representative datasets have been introduced in recent years. These datasets include videos from sports , movies, daily lives and person-environment interaction videos \cite{gu2018ava,sigurdsson2016hollywood,abu2016youtube,monfort2019moments,lee2012discovering,Damen2018EPICKITCHENS,caba2015activitynet,cvpr18waqas,jiang2014thumos}.

In what follows, we provide a brief review of some of the very important visual features techniques, machine learning approaches to learn action classifiers and some of the recent action datasets.

\section{Visual Features for Action Recognition}
To recognize and localize human action in videos, several recent visual features have been proposed. Good visual features are invariant to scale, rotation, affine transformation, brightness changes, occlusion and camera motion, and position. Overall, there are two types of features, i.e., handcrafted features and features learned through deep networks. In handcrafted features, there are further two categories. Local features are extracted by the dense sampling of the videos or finding interest points in frames whereas holistic features gather features that extract global shape, structure and contextual information of the human body and make a 3D volume of space-time. These features contain the human pose information at a different time and spatial location of the person in video frames. Deeply learned features capture both local and global information in the same framework.
\subsection*{Hand-Crafted Visual Features}
\textbf{Holistic features:} Holistic representations extract features from global regions (whole frame or whole human body) which are invariant to the cluttered background and appearance changes.   
Yilmaz et al. \cite{yilmaz2005actions} purpose space-time volume (STV) to take space-time information of action. SVT is generated by stacking the 2D object contour in the image plane with respect to time to make a space-time volume.
Differential geometric properties from STV are shown to be the invariant viewpoint.
Another technique of getting motion information is through algorithm optical flow which computes the direction of motion on two consecutive frames.  The shape model presented in \cite{jiang2012recognizing} learned a prototype tree for action recognition.

\noindent\textbf{Local Features:}
In local representation, Spatio-temporal keypoints (corners, edges, etc) are detected in the video and descriptors made over these key points are used to capture the local motion information. Laptev et al. \cite{laptev2005space} (called STIP) extended Harris corner detector in space-time domain. They used a normalized Spatio-temporal Laplacian operator to detected events over temporal and spatial scales. Local representation overcomes the problems in holistic representation.

Spatio-temporal interest points capture information for a short duration of time and hence cannot capture long-term duration information.  Wang et al. \cite{wang2011action} extract dense trajectory features for capturing long-duration information. Feature points are densely extracted on gird of pixels and these points are tracked in consecutive frames to make dense trajectory. HOG (histograms of oriented gradients), HOF (histograms of optical flow) and MBH (Motion Boundary Histogram) are computed along the trajectory to extract static appearance information, local motion information and encode relative motion information respectively. This method is shown to perform much better than STIP because trajectory captures the motion and dynamic information. 
\subsection*{Deep Network-based Features}
With the resurgence of deep learning, several new features are introduced for action recognition. Below, we briefly explain some of the most used deep features.\newline
\textbf{Two-Stream Convolutional Networks:}
Simonyan et al.\cite{simonyan2014two} proposed two-stream deep networks for action recognition that have two separate input streams (spatial, temporal). The spatial stream uses information from still video frames while the temporal stream uses the dense optical flow. Both streamed are fused to produce the final output. Their model is inspired by the two-streams hypothesis in which the human visual cortex contains two paths. The ventral stream (which performs object recognition) and the dorsal stream (which recognizes motion).\newline
\textbf{3D ConvNets (C3D):} Tran et al. \cite{tran2015learning} purpose deep 3-dimensional convolutional networks (3D ConvNets) for learning of spatiotemporal features and a simple linear classifier. They showed good performance on different video analysis tasks with these learner features. Network architecture is as follow, it has 8 convolution layers, 5 pooling layers, followed by two fully connected layers, and a softmax output layer. Convolution kernels are of size 3 $\times$ 3$\times$ 3 whereas the pooing kernels are of size 2 $\times$ 2$\times$ 2 except for first pooling layer which has 1 $\times$ 2$\times$ 2. Finally, each fully connected layer has 4096 output units.\newline
\textbf{Inception-3D:} Carreira et al. \cite{carreira2017quo} purpose a new two-stream inflated 3D ConvNet called “I3D” in which 2D ConvNet trained for image classification is expanded to 3D ConvNet that extract spatio-temporal feature from a video. They convert 2D ConvNets that accurately work with 2D-image classification models into a 3D model by adding one additional dimension to 2D filter and kernel. Resulted kernel and filters have an additional temporal dimension where 2D - N x N filters converted into 3D - N x N x N. As mentioned before, inspired by two-stream networks \cite{simonyan2014two} for videos classification, Carreira et al. \cite{carreira2017quo}  use two 3D-Streams; one for RGB and other for optical flow. Both networks are trained separately and the results are averaged. \newline 
\textbf{Multi-Fiber Networks} Chen et al. \cite{chen2018multi} purposed multi-fiber networks architecture that composed of separately connected multiple fibers or lightweight 3D convolutional networks which are independent of each other. In this way, a complicated neural network is divided into a group of different small networks. They increase the model efficiency by reducing the number of connections in the network. The number of connections is reduced by slicing the conventional complex residual unit into fixed separate parallel paths (called fibers). They solve the information blockage problem across the paths using fully convolution layer at the beginning and end of each unit and use a multiplexer that redirects and amplifies features from all fibers. The paper shows state-of-the-art action recognition accuracy with less computational time.

\section{Action Datasets}
In this section, we briefly review some of the recent action detection datasets.

\subsection*{THUMOS-14}
THOMAS 14 \cite{jiang2014thumos} contains videos of a large number of human action classes. The dataset contains a variety of actions including normal daily life activity (brushing teeth) and sports actions (Golf Swing). It contains a total of 18,394 video sequences.
They have used entire UCF-101 data for training and testing is done on 1579 videos that contain one or more action instances in it.

\subsection*{UCF-Crime}
Sultani et al. \cite{cvpr18waqas} proposed a new anomaly detection dataset named as UCF-Crime. The UCF-Crime dataset contains untrimmed real-world surveillance videos. It contains 13 real-world anomalies, including Abuse, Arrest, Arson, Assault, Road Accident, Burglary, Explosion, Fighting, Robbery, Shooting, Stealing, Shoplifting, and Vandalism. This dataset contains a total of 1900 video clips. The total duration of 1900 clips is 128 hours at 30 fps with 240 $\times$ 320 resolutions. They use 800 normal and 810 anomalous videos for training and the remaining 150 normal and 140 anomalous videos are used for testing. 

\subsection*{ActivityNet}
ActivityNet \cite{caba2015activitynet} is a large scale action dataset that contains 203 activity classes like eating, drinking, sport, exercise and relaxing. This dataset contains a total of 1900 video clips. The total duration of 19,994 clips is 849 hours at 30 fps with 1280 $\times$720 resolutions. They have used 50\% videos for training, 25\% videos for testing and the remaining 25\%  videos for validation. 

\subsection*{EPIC-Kitchens}
The epic-kitchens dataset \cite{Damen2018EPICKITCHENS} contains videos of different kitchen actions. In these videos, the camera is mounted on the face of the person. Videos are taken from 32 different kitchens. The dataset contains 819 different actions like a wash, adjust heat, pour oil and put the bottle. The total duration of video clips is 55 hours having 45 million frames at 60 fps and 1920 $\times$ 1080 resolutions. They split the dataset in which 80\% videos are used for training and the remaining 20\% are used for testing.

\subsection*{UT Egocentric}
UT Egocentric dataset \cite{lee2012discovering} contains four videos of different action captured by a camera mounted on the head of a person.  The duration of these videos are about 3 - 5 hours long and are captured in a natural and uncontrolled environment. The actions include driving, eating, shopping, cooking and attending lectures. The total duration of 4 clips is 17 hours at 15 fps with 320 $\times$ 480 resolutions. In this dataset 3 videos are used for training and 1 video is used for testing.

\subsection*{Moments in Time}
Moments in Time dataset \cite{monfort2019moments} is generated by MIT-IBM Watson AI Lab to help the vision system to understand and recognize the action in the videos. The dataset contains videos of people, animals, objects or natural phenomena. It contains 339 action classes and one million labeled 3-second videos at 5fps. They generate a training set of 802,264 videos having 500 to 5000 videos per class. The test set of 67,800 videos has 200 videos per class. 
\subsection*{YouTube M8}
YouTube M8 \cite{abu2016youtube} was generated by Google in 2016 which contains 6.1 million video ID taken from YouTube and duration of video clips are 350,000 hours at 1fps. There are approximately 3862 classes in this dataset and class labels are machine-generated. In this dataset, a single video has multiple labels and average labels per video are around 3. The dataset contains pre-computed audio and visual features from video frames that are easily used to train machine learning models. Both feature and video level labels are available for download.
They split videos into 3 partitions, Train 70\%, Validate 20\% and Test 10\%. They want to make their dataset as a baseline to evaluate the various classification models using popular evaluation metrics.
 
 \subsection*{Charades}
 Charades \cite{sigurdsson2016hollywood} is a very large temporal video action dataset that is presented in ECCV 2016. To make their dataset more realistic they record different indoor action videos by 267 unique users according to predefined sentences. These sentences are made from fixed vocabulary which includes objects and actions of different kinds. The dataset contains a total of 9848 annotated videos of 157 different actions having a length of 30s. The annotation contains both descriptions of the video and temporal intervals of different performing actions.

 \subsection*{AVA Actions Dataset}
 Atomic Visual Actions (AVA) \cite{gu2018ava} is a densely labeled video action dataset of untrimmed videos. In this dataset, actions are labeled with respect to both temporal and spatial locality and have 80 different atomic visual actions. Multiple labels on single humans are annotated, resulting in 1.62M of total dens labels. It contains 430 different video clips having a duration of 15 minutes. Every person is labeled with a bounding box and type of action from its atomic visual action vocabulary. 

\begin{figure*}[!h]
    \centering
    \includegraphics[width=1\textwidth]{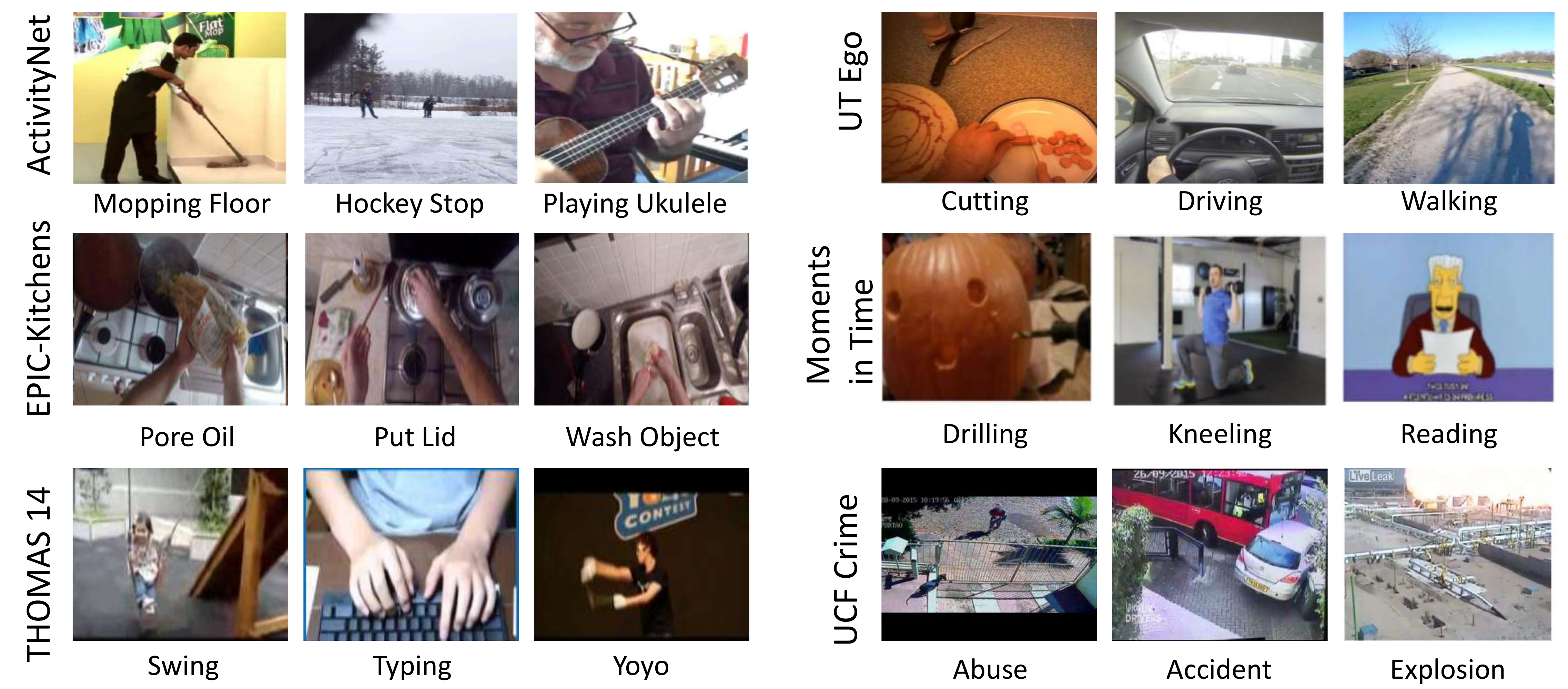}
    \caption{Typical frames from some of the most popular action datasets}
    \label{fig:my_label}
\end{figure*}

\section{Action Recognition Approaches}
In what follows, we briefly review some of the recent fully supervised, weakly supervised and unsupervised action recognition approaches.

\subsection*{Supervised Action Recognition}
Supervised action recognition assumes the availability of complete labels of all the classes to be learned. Although, getting videos annotated is quite a time consuming and costly task, the supervised action recognition methods have much higher detection accuracy as compared to unsupervised or weakly supervised approaches.

Jain et al. \cite{jain201515} presented that object encoding improves the performance of action localization and classification. They have obtained responses of 15,000 object detectors for each frame of video and averaged those responses to video representation. They demonstrated that these object-based representations provide good action recognition and detection accuracy. They also demonstrated that object-action relations are generic and it can be used for transfer learning between different datasets. 

Choutas et al. \cite{Choutas_2018_CVPR} proposed an interesting approach for robust human recognition using a new representation called potion. The potion is obtained by temporally aggregating the probability
maps of the human pose estimator. They assigned different colors to human poses depending on their relative temporal location in the videos.

Recently, Hu et al. \cite{hou2019efficient} proposed an encoder-decoder framework using 3D separable convolution for the pyramid pooling of efficient human action recognition and segmentation. Their proposed approach both efficient and provides very good segmentation and action detection as compared to several competitive baselines.

\subsection*{Weakly Supervised Action Detection}
Weakly supervised action detection fall between supervised learning (labeled data) and unsupervised learning (no labels). In weakly supervised learning, the complete annotations of the concepts to be learned are not available during training. For example, for spatio-temporal action detection, instead of spatio-temporal bounding boxes, only the video level labels are available during training. However, on testing, the classifier needs to provide spatio-temporal bounding boxes of actions. Weakly supervised approaches help in reducing the time, effort and cost of annotations.

Nguyen et al. \cite{cvf18nguyen} proposed a sparse pooling network to temporally localize human action in the videos. They introduced a method to generate temporal class activation mapping in the two-stream framework and demonstrated improved detection results.

Wang et al. \cite{cvpr17wang} introduced a new end to end architecture, called UntrimmedNet. It generates short clips proposal from videos by uniformly sampling. After extracting the network from a pre-trained network, action labels are predicted for each temporal segment. Furthermore, the selection module is proposed to rank important action proposals. Finally, the output of the classification and selection module are fused to produce a final classification.

Singh et al. \cite{kumar2017hide} used a different and new approach to that problem called ‘Hide and Seek’. Instead of changing the algorithm, they change the input video. During training, they randomly remove the frames and hence force the network to learn all the discriminative frames which produce good classification results. These automatically discovered frames are then used for action classification.

Chang et al. \cite{chang2019d} purposed Discriminative Differentiable Dynamic Time Warping (DTW) that uses weak supervision to segments and aligns the video frames. At training time, only the ordered list of action is provided. The main contribution of their work is to make an alignment loss to be differentiable.

Weakly supervised anomaly detection algorithms in \cite{sultani2018real} developed multiple instance ranking loss for criminal activity detection in surveillance videos. Training labels of being normal and abnormal are assigned at video level and a model is a train to temporally detect abnormal activists in videos.

\subsection*{Unsupervised Action Recognition}
Due to the availability of free humongous visual data, several researchers have worked on designing unsupervised approaches for action recognition. The key advantage of unsupervised action recognition is that it does not need any kind of manual annotations. The design of such unsupervised methods can save time and cost and evade manual annotations biases.  

Generally, in unsupervised action recognition, the first step is clustering. The visual features are extracted from the videos and based on feature similarities, videos are grouped into separate clusters. After that, the classifier is learned on these clusters. This is followed by the iterative process in which both classifiers and clusters are improved. As compared to fully supervised action recognition, less research work has been done in unsupervised action recognition. Below, we describe some of the recent works for unsupervised action recognition.

Jones et al. \cite{jones2014multigraph} proposed feature grouped spectral multigraph (FGSM) approach for unsupervised action recognition.  Firstly, feature clustering generates the number of action classes or feature space.  Secondly, for each feature, a separate graph is generated. Multigraph Spectral Embedding is found on each graph and these embeddings are combined into a single representation.

Yu et al. \cite{yu2011unsupervised} introduced a technique in which video frames are categories by a collection of spatio-temporal interest points (STIPs). They use the histogram of gradient (HOG) and histogram of flow (HOF) as a descriptor of STIP. A random forest is built to model the distribution of high-dimensional feature space. After that, from this random forest, each STIP is matched to the query class and provide a voting score for each action type. 

Recently, Soomro et al. \cite{soomro2017unsupervised} proposed a novel unsupervised approach for action detection. They clustered action videos using a dominant set clustering algorithm. After that discriminative clustering is applied, followed by the variant of Knapsack optimization. They have demonstrated competitive results compared to several baselines. Finally, Sultani et al. \cite{sultani2017unsupervised} proposed an unsupervised way to rank the action proposal in the videos. They demonstrated that better re-ranking of action proposals leads to better action detection accuracy.
 
 \begin{figure*}[!h]
    \centering
    \includegraphics[width=1\textwidth]{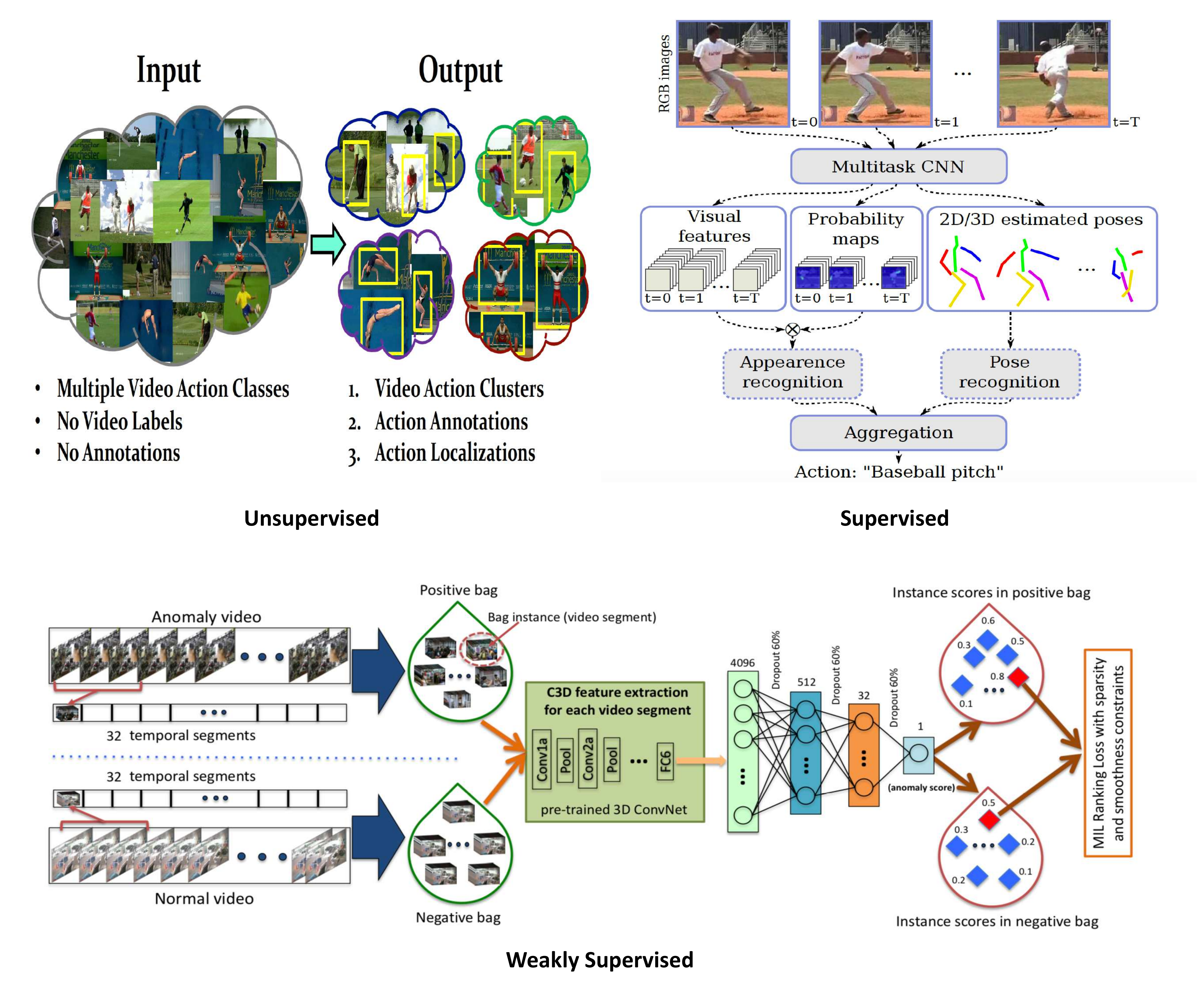}
    \caption{Examples of recently proposed Unsupervised \cite{soomro2017unsupervised}, Weakly Supervised \cite{cvpr16waqas} and Supervised \cite{Supervised2D3D} action classification approaches. }
    \label{fig:my_label}
\end{figure*}
 
\subsection*{Action Segmentation}

The desired output of the action segmentation is to generate a segmentation mask on the regions of video where the action is being performed including start time, end time, and action class label. Many papers use different techniques for action segmentation. Jhuang et al. \cite{jhuang2013towards} describe the importance of action segmentation in video data. They show that different types of annotations on the same dataset change the performance of the algorithm. They described that 2D puppet model of humans (made from 10 body parts connected by 13 joints) show significant improvement in classification rather than using bounding boxes, class level label or temporal localization of action in the video. Lu et al. \cite{lu2015human} found a single and hierarchical MRF model which showed significantly improved results in action segmentation. They also found that bounding box and action segmentation mask improves results over the video level label and specifically segmentation mask gets better results than bounding box.  
Ghosh et al. \cite{ghosh2020stacked}  introduced stacked spatio-temporal graph CN for action segmentation. Graph-based CNN uses a combination of spatial and temporal dynamics for better segmentation. Their method is an extension of the spatio-temporal graph CN that was developed for skeleton-based action recognition.

\subsection*{Trimmed video action classification}

Trimmed action videos contain single action from start to end of the video. These types of videos are usually shorter in length around 10 - 15 sec. 
Sultani et al. \cite{sultani2016if} introduced a novel approach in which they used web images to achieve spatial action localization in trimmed action videos.  
Soomro et al. \cite{soomro2015action} used supervoxels for capturing the relation between spatio-temporal segmentation in the video. Initially, they started with random supervoxels during training and then found matching supervoxels and used those to localize action at testing time.  
Carreira et al. \cite{carreira2017quo} purpose a two-stream Inflated 3D ConvNet called “I3D” that extracts spatial-temporal feature form video. The idea behind I3D is to expand 2D ConvNet trained for image classification into 3D ConvNet for video action classification. They convert 2D ConvNets that accurately work with 2D-image classification models into a 3D model by adding one additional dimension to the 2D filter and kernel. 

\subsection*{Untrimmed video classification}

Untrimmed videos are comparatively longer in time, contain multiple actions and/or single action is repeated several times in the same video. They are more close to real-world settings. These types of videos are usually around 5 to 15 minutes long.  
Singh et al. \cite{singh2016untrimmed}  introduced a method in which they performed the tasks of action classification by combining video level global feature and frame-level features and generated temporal action proposals by using dynamic programming. Montes et al. \cite{montes2016temporal} introduced a  simple method in which they used the features from 3D Convolutional Neural Network (C3D) and employ recurrent neural networks (RNN) to perform action classification.

\subsection*{Temporal Action Detection}
Temporal action detection algorithms seek to identify the frames of the video where the action is being performed. Recently, several approaches are being presented for the temporal action detection including \cite{singh2016untrimmed,montes2016temporal,zhao2017temporal}. Zhao et al. \cite{zhao2017temporal} purposed a novel approach called a structured segment network (SSN) in which the temporal structure of action is represented with a structured temporal pyramid.

\subsection*{Spatio-Temporal Action Detection}
Spatio-temporal action classification is a much harder problem where we want to do temporal localization (starting and ending frame of the action) as well as spatial localization.%
Several of the above mentioned papers accomplished spatio-temporal action classification \cite{sultani2016if,singh2017online}. Singh et al. \cite{singh2017online} introduced a new efficient action tube generation algorithm employing optical flow for action localization and classification.

\section{Action Recognition Challenges}
 Action Recognition Challenges aim to mature the computer vision algorithm across different domains such as surveillance videos, indoor/outdoor activities, wildlife observation and action Anticipation 
 Generally, the organizing committees of these challenges release some new large scale datasets and participants across all over the world compete with each other on these datasets. Below are some of the famous computer vision competitions.
 \subsection*{THUMOS Challenge}
 THUMOS workshop and challenge \footnote{http://www.thumos.info/home.html} contributes a lot in defining new challenges and approaches in automatic action recognition and localization. The challenge is performed on the THUMOS 2015 dataset which is a very large action recognition dataset of untrimmed videos recorded in realistic scenes taken from youtube.  THUMOS 2015 is an extension of THUMOS’14 but it has 430 hours of videos that are 70\% larger than THUMOS’14.  The participants trained their models and checked their performance on both action classification and temporal action localization.

 \subsection*{AVA Challenge}
 AVA Action challenge \footnote{https://research.google.com/ava/challenge.html} aims at exploring new approaches for action recognition in both space and time on the AVA dataset. In this challenge, participants have to identify 80 video classes of actions. Performance is evaluated on the localization of action in both time and space. AVA dataset is quite challenging as multiple people are doing multiple actions in the video. In 2019 Facebook Al Research (FAIR) won this challenge with 34.24\% mAP at 0.5 IOU.
 
 \subsection*{EPIC-Kitchens Action Recognition challenge}

EPIC-Kitchens Action Recognition challenge \footnote{https://epic-kitchens.github.io/2020} aims to classify trimmed videos of seen and unseen kitchens action in EPIC-Kitchens dataset. Videos are recorded by a camera mounted on the face of people in 32 different kitchens. The task in this challenge is to classify the action segments from a trimmed video. Noun and verb classes jointly define the action class for a segment. Participants have to provide confidence scores for each noun and verb class for their submission. The highest top-1 accuracy of this challenge in 2019 on the action recognition task is 41.37\% with 25.13\% precision and recall 26.39\%.

 \subsection*{Charades Activity Challenge}
 Charades Activity Challenge \footnote{http://vuchallenge.org/charades.html} aims to explore challenges and methods on automatic action recognition tasks on indoor daily life activities of people, by providing realistic videos from Charades dataset. Charades is a very large dataset of diverse videos from daily life activities including sitting on a chair, opening doors, working on computers and drinking water. The aim is to boost the action recognition accuracy in real daily life tasks. There are two separate tracks in this challenge: classification track and localization track. The classification track is to classify all activities/actions of the given video. The localization track is to localize the intervals of a specific action. The top accuracy of the winning team in this challenge is 34\% mAP.

\section{Open problems}
Although tremendous research work has been done in different areas of action recognition problems, there are still several areas that are less explored. CCTV surveillance cameras are ubiquitous nowadays and recording humongous about of data 24/7. Most of these videos are of low quality. There is not much research work for the detection of actions in CCTV cameras. Stat of the art action recognition methods performed quite poorly on CCTV videos. Action detection under different weather conditions (rain, snow, shadow) is not much explored. Action detection in the videos capturing night scenes is still an unexplored area.  Only a little research work has been done for action recognition from far cameras (cameras mounted to the top of the building). Furthermore, a lot of work needs to be done for action detection and localization in crowded environments.



\bibliographystyle{plain}
\bibliography{template}
\end{document}